# Adaptation and Fine-tuning with TabPFN for Travelling Salesman Problem

Nguyen Gia Hien Vu[1], Yifan Tang[1], Rey Lim[2], Yifan Yang[3], Hang Ma[3], Ke Wang[1], and G. Gary Wang[1,4]

**ABSTRACT**

Tabular Prior-Data Fitted Network (TabPFN) is a foundation model designed for small to medium-sized tabular data, which has attracted much attention recently. This paper investigates the application of TabPFN in Combinatorial Optimization (CO) problems. The aim is to lessen challenges in time and data-intensive training requirements often observed in using traditional methods including exact and heuristic algorithms, Machine Learning (ML)-based models, to solve CO problems. Proposing possibly the first ever application of TabPFN for such a purpose, we adapt and fine-tune the TabPFN model to solve the Travelling Salesman Problem (TSP), one of the most well-known CO problems. Specifically, we adopt the node-based approach and the node-predicting adaptation strategy to construct the entire TSP route. Our evaluation with varying instance sizes confirms that TabPFN requires minimal training, adapts to TSP using a single sample, performs better generalization across varying TSP instance sizes, and reduces performance degradation. Furthermore, the training process with adaptation and fine-tuning is completed within minutes. The methodology leads to strong solution quality even without post-processing and achieves performance comparable to other models with post-processing refinement. Our findings suggest that the TabPFN model is a promising approach to solve structured and CO problems efficiently under training resource constraints and rapid deployment requirements.

## 1. INTRODUCTION

Combinatorial Optimization (CO) is a branch of mathematical optimization that aims at optimizing discrete variables with nonpolynomial increase of possible solutions with the given number of variables (Cappart et al., 2023). CO problems are hence considered NP-hard (Cappart et al., 2023). In real world, CO problems are commonly seen both in research and industries, such as logistics, production scheduling, or transportation (Qiu et al., 2022; Sun & Yang, 2023; Li et al., 2024). The Travelling Salesman Problem (TSP) is one of the most well-known problems in the CO domain (Alanzi & Menai, 2025; Saller et al., 2025), arising from routing and scheduling applications (Jaillet & Lu, 2011). Conceptually, TSP is a graph-based problem defined by a set of nodes, and the goal is to visit each node exactly once, then return to the starting point, with the minimum total route cost such as route length or time (Vesselinova et al., 2020). TSP has been studied in thousands of publications each year (Saller et al., 2025), as well as in a wide range of applications such as those in engineering, biology, or economics, just to name a few (Johnson & Liu, 2006; Crişan et al., 2020; Singh et al., 2022).

As reviewed by Alanzi & Menai (2025), two traditional approaches to deal with TSP are exact and heuristic algorithms. Exact algorithms are designed to make sure that the global optimal is found for each sample (Applegate et al., 2003a). However, it might become impractical to find an optimal solution in a reasonable amount of time when the instance size grows. Even, this approach can take years to solve large-sized cases (Applegate et al., 2003a). As improvements, heuristics and metaheuristics are developed to improve computational efficiency (Rego et al., 2011). Notably, they are effective on small to medium-sized TSP problems. However, similar to exact algorithms, heuristics and metaheuristics might fail to return optimal solutions within reasonable time frames, especially for new TSP variants (Rego et al., 2011). Therefore, there should be a better TSP-solving algorithm that can both meet timeframe requirements and maintain the quality of the final solution.

---

[1] Product Design and Optimization Laboratory, Simon Fraser University, Surrey, BC, Canada
[2] Simon Fraser University Alumnus, Burnaby, BC, Canada
[3] School of Computing Science, Simon Fraser University, Burnaby, BC, Canada
[4] Corresponding author: Email gary_wang@sfu.ca Tel: 778 782 8495

In this trend, Machine Learning (ML), especially Deep Learning (DL) models, gains significant attention in recent years (Kool et al., 2019; Kwon et al., 2020; Qiu et al., 2022; Sun & Yang, 2023; Li et al., 2024; Meng et al., 2025; Zhao & Wong, 2025). Different research and publications evidence the success of this approach in solving TSP problems rapidly once the training has been completed without sacrificing the quality of the solution, revealing considerable potential to apply ML techniques such as transformers, attention mechanisms, or diffusion models (Kool et al., 2019; Kwon et al., 2020; Qiu et al., 2022; Sun & Yang, 2023; Li et al., 2024; Meng et al., 2025; Zhao & Wong, 2025). However, this trend often meets four barriers as follows:

- All these models require extremely long training time, as well as a dedicated specialized GPU (Kool et al., 2019; Kwon et al., 2020; Qiu et al., 2022; Sun & Yang, 2023; Li et al., 2024; Meng et al., 2025; Zhao & Wong, 2025). For example, to train their specialized diffusion model, Li et al. (2024) needs over 20 days using a specialized A100 GPU. Similarly, according to Kwon et al. (2020), their training time can take up to one week for a full convergence.
- Many current models employ probabilistic approaches and often become untrainable on large-sized TSP problems that have more than 100 nodes (Sun & Yang, 2023). Some researchers have attempted training their models on larger instance sizes, but each of the varying TSP instance sizes may require a respective specialized model, or else the performance could suffer (Joshi et al., 2021; Li et al., 2024).
- A large quantity of training samples is required. For instance, Kwon et al. (2020) generate up to 200 million samples for the whole training process, while the model by Zhao & Wong (2025) requires over one million samples.
- Much of the current research encodes the entire problem, i.e., whole-instance encoding, regardless of the input instance size, as seen in the model by Zhao & Wong (2025). This limits the efficiency of encoding TSPs of varying sizes with different characteristics, particularly when the ML model size is intended to remain relatively constant.

In this context, foundation models have emerged as a potential candidate to alleviate or resolve those barriers. As large-scale Artificial Intelligence (AI) models which are pre-trained on a vast amount of data, foundation models exhibit strong capabilities from different perspectives, such as those in creative generation, software debugging, protein sequencing, and text-to-image tasks (Schneider et al., 2024). As these models scale, they become better at performing tasks beyond their original training, broadening their range of applications without requiring additional training or fine-tuning. Furthermore, when task-specific performance is needed, fine-tuning or prompt engineering can enhance their effectiveness at significantly lower costs, both in terms of time and data, compared to developing entirely new models (Schneider et al., 2024). Inferably, improvements in data efficiency and rapid adaptation can be expected when foundation models are used. However, as observed by Tu et al. (2024), foundation models often have a high number of parameters, resulting in intensive calculation during inference, and their sizes likely reach up to billions of parameters.

Among foundational models, Tabular Prior-data Fitted Network (TabPFN) is a newly designed model for small to medium-sized tabular data, requiring a small training data size and short training time (Hollmann et al., 2023; Hollmann et al., 2025). As the latest version of TabPFN, TabPFN-v2 is advantageously seen outperforming current leading models by a substantial margin across various benchmark tests (Hollmann et al., 2025). Distinctly, although they possess such prominent advantages, TabPFN in general and TabPFN-v2 in particular have considerably fewer parameters compared to other existing foundation models (Hollmann et al., 2023; Hoo et al., 2025). In reality, TabPFN-v2 has seen its applications in different fields, such as time series forecasting (Hoo et al., 2025), geotechnical (Saito et al., 2025), and medical applications (Luo et al., 2025). Notably, TabPFN-v2 has not been employed in tackling CO problems yet.

The paper explores the applicability of TabPFN-v2 in CO problems out of which TSP is chosen as a representative. We adapt and fine-tune TabPFN-v2 using a single sample specific to TSP, adopting the node-based approach, with the goal of minimizing both data requirements and training time for varying TSP instance sizes for improved data efficiency and rapid adaptation. Specifically, the model is adapted and fine-tuned to predict the location of the next node, which can then be used to construct the entire TSP route. Notably, this node-based approach is less prone to expressiveness issues than whole-instance encoding methods which require different model training for varying instance sizes, such as the case proposed by Li et al. (2024). Aiming to show the potential of TabPFN-v2 in solving varied-sized problems without the need for extensive retraining or fine-tuning, this paper is targeted at four key contributions as follows:

- The first application of TabPFN-v2 to CO problems: To the best of the authors' knowledge, this paper presents the first ever application of TabPFN-v2, a supervised learning model, to CO problems.
- The node-based encoding approach for better scalability: Our node-based approach can avoid expressiveness issues inherent in whole-instance encoding methods, such as those used by Li et al. (2024). This allows our model to better handle larger TSP problems without the need for extensive retraining or fine-tuning, thus making it a scalable solution for real-world applications.
- An efficient node-predicting adaptation strategy for TSP: We introduce a novel adaptation method that allows TabPFN-v2 to predict the next node in the TSP route efficiently, using a single sample for this process. This strategy aims to minimize training time, hardware resources, and data requirements, thereby offering a faster and more resource-efficient solution compared to traditional methods.
- Benchmarking and evaluation results: We conduct comprehensive experiments using our adapted and fine-tuned model (hereinafter referred to as the adapted model) on varying TSP instance sizes, comparing the performance of our TabPFN-v2-based model against state-of-the-art (SOTA) algorithms. The results highlight competitive edges of our adapted model in terms of both solution quality and computational efficiency.

Within such a scope, the remainder of this paper is structured as follows: Section 2 briefly reviews TSP, TSP solving approaches, which include traditional ones such as exact algorithms, heuristics, and ML, and provides a brief overview of foundation models. Section 3 is dedicated to TabPFN with a particular focus on TabPFN-v2. Section 4 discusses our proposed method and experimental setup, covering data preparation, training and solution decoding, post-processing, and benchmarking and evaluation scheme. Section 5 presents and discusses our experiment results. Finally, Section 6 summarizes the findings and recommends possible future research.

## 2. LITERATURE REVIEW

### 2.1 Travelling Salesman Problem (TSP)

*2.1.1 Problem Definition and Formulation*

As briefly mentioned in Section 1 of the paper, TSP is a problem involving a set of nodes, where the goal is to determine the lowest-cost route that visits each node exactly once and returns to the starting node ($c$) (Laporte, 1992; Vesselinova et al., 2020). In formal terms, TSP is defined on an undirected, edge-directed graph $G = (V, E)$, where $V$ represents the set of $n$ nodes and $E$ is the set of edges connecting those nodes ($E = \{e_{ij}\}$ for $(i, j = 1,2,3, \ldots, n; i \neq j)$). Each edge has a corresponding cost $c_{ij}$. The set of feasible solutions $F$ includes all edge subsets that form a Hamilton cycle, and the objective function to minimize is the sum of edge weights $w(e)$ for all edges in the solution $f \in F$ (Laporte, 1992; Vesselinova et al., 2020). TSP can be formulated using binary variables $x_{ij}$ where $x_{ij} = 1$ if edge $e_{ij}$ is used in the optimal solution $(i \neq j)$, or else $x_{ij} = 0$. Altogether, TSP formulation can be stated as follows (Laporte, 1992):

$$\text{Minimize } w(e) = \sum_{i \neq j} c_{ij} * x_{ij} \quad \text{(Equation 1)}$$

S.T.

$$\sum_{i=1}^{n} x_{ij} = 1, \forall j \in V \quad \text{(Equation 2)}$$

$$\sum_{j=1}^{n} x_{ij} = 1 \ \forall i \in V \quad \text{(Equation 3)}$$

$$\sum_{i,j \in S} x_{ij} \leq |S| - 1 \ \forall S \subset V, 2 \leq |S| \leq n - 2 \quad \text{(Equation 4)}$$

$$x_{ij} \in \{0, 1\}, \forall i, j \in V, i \neq j \quad \text{(Equation 5)}$$

In this formulation, the objective (Equation 1) describes the cost of the optimal tour. Equations 2 and 3 guarantee the flow of the route, which means each node should be entered exactly once and left exactly once. Equation 4 eliminates a potential subtour, i.e., a full Hamilton cycle on a subset that does not include all $n$ nodes. Finally, Equation 5 imposes a binary constraint on decision variables (Laporte, 1992).

Historically, TSP is a relatively old problem and has been mentioned as early as the 18[th] century (Larrañaga et al., 1999). Incrementally, as observed by Larrañaga et al. (1999), TSP has occupied the interest of many researchers for different reasons, including:

- TSP is easy to describe, yet a hard problem, especially due to the lack of an NP-complete algorithm.
- TSP has broad applicability which goes beyond the normal daily routing, as it is also applicable in areas such as scheduling, planning, or even cryptoanalysis.
- TSP has become a test problem, especially to test the usefulness of a new CO method.

Throughout the years, more specialized variants have also been researched to meet different requirements, including Symmetric TSP (STSP) (Carpaneto et al., 1989), Asymmetric TSP (ATSP) (Frieze et al., 1995), and Multiple TSP (MTSP) (Bektas, 2006). Regardless of the variant in question, TSP solving approaches can generally be categorized into traditional methods and modern ML-based ones, which are discussed in the subsequent Subsections 2.1.2 and 2.1.3, respectively.

*2.1.2 Traditional TSP Solving Methods*

Traditional algorithms for TSP include two main branches, namely exact algorithms and heuristic algorithms (Alanzi & Menai, 2025). Exact algorithms try to find optimal solutions by exploring the entire solution space exhaustively (Alanzi & Menai, 2025).

There are different approaches for exact algorithms, such as Dynamic Programming (DP) or Branch-and-Bound (BB). DP approach, such as the Held & Karp (1962) algorithms, tries to break TSP problems into smaller subproblems to avoid redundant computation. Although this approach ensures the optimality of the final solution, it requires exponential time complexity and high resource requirements, resulting in impracticality for large-size TSP problems, as illustrated by Yang et al. (2023). Different from the DP approach, the BB approach, as illustrated by Balas & Toth (1983) algorithm, explores the solution space using branching to search the corresponding space and eliminate any suboptimal branches. However, similar to the DP approach, this approach still has relatively high time complexity (Alanzi & Menai, 2025). Furthermore, Concorde (Applegate et al., 2003b), an advanced exact TSP solver frequently used as a benchmark in various studies (Kool et al., 2019; Kwon et al., 2020; Qiu et al., 2022; Sun & Yang, 2023; Li et al., 2024; Meng et al., 2025; Zhao & Wong, 2025), continues to struggle with large-size TSP problems, as highlighted by Lu et al. (2022).

To resolve time complexity issues, heuristic and metaheuristic approaches have been developed. One of the most famous SOTA solvers in this category is the LKH algorithm (Helsgaun, 2017), which employs k-opt strategies to conduct a local search process. Notably, while heuristic approaches are significantly faster than exact algorithms, they are developed for specific TSP problems. For instance, in Helsgaun (2017) 's report, only a limited number of solvers are introduced for specific TSP variants, which makes it difficult to adapt and fine-tune the approach when new constraints are introduced. Furthermore, for large-size TSP problems, the search process might still take hours on a multicore CPU (Skinderowicz, 2022).

*2.1.3 Machine Learning Application*

In recent years, AI and ML have attracted much attention (Hajkowicz et al., 2023). In the CO domain, especially TSP, we see a considerable number of relevant publications (Kool et al., 2019; Kwon et al., 2020; Qiu et al., 2022; Sun & Yang, 2023; Li et al., 2024; Meng et al., 2025; Zhao & Wong, 2025). According to Alanzi & Menai (2025), ML-based approaches can be categorized into traditional ML-based approaches and DL-based approaches.

In ML-based approaches, traditional TSP algorithms can be combined with different ML models to improve the search process. For example, Zheng et al. (2023) combine reinforcement learning (RL) with the LKH solver to improve traversal operations. The results from the paper show that it can outperform current SOTA heuristics on certain problems (Zheng et al., 2023). In DL-based approaches, different neural networks, such as Graph Convolution Networks (GCN) (Joshi et al., 2019), diffusion models (Sun & Yang, 2023), or transformers (Zhao & Wong, 2025) are utilised to capture TSP characteristics and optimize the solution (Alanzi & Menai, 2025). Different results, such as the findings by Li et al. (2024), have shown that they can achieve promising outcomes in a relatively short amount of time compared to traditional metaheuristic solvers.

Despite such potential, these approaches still reveal multiple challenges. For example, as briefly discussed in Section 1, although recent advancements have expanded their capacity to handle larger samples, training time remains considerably prohibitive (Kwon et al., 2020; Li et al., 2024; Alanzi & Menai, 2025). This is likely due to slow convergence and labeled data inadequacy for many models to perform well (Alanzi & Menai, 2025). Besides that, there are concerns about their ability to get generalized to unseen TSP problems of different sizes, their requirements for manual tuning, or their risks of performance breakdown when a large number of agents in a RL-based method are employed (Alanzi & Menai, 2025).

### 2.2 Foundation Models

Foundation models can be defined as extensive AI systems which are pre-trained on massive datasets of general information with different learning techniques and adaptable to various downstream tasks (Bommasani et al., 2022). Foundation models often have a large number of parameters, such as LLaMA with up to 65 billion parameters (Touvron et al., 2023), TimesFM with 200 million parameters (Das et al., 2024), or GPT-3 with 175 billion parameters (Brown et al., 2020). These models can be customized for particular tasks by further adapting them to domain-specific data, such as fine-tuning the pre-trained models (Schneider et al., 2024). In other words, this is similar to transfer learning where the foundational knowledge learned during the pre-training is transferred to new tasks with much smaller amounts of data (Zhou et al., 2024). This enables the model to use the domain-specific data more efficiently and adapt to these new tasks faster compared to training a model from scratch. In fact, many different foundation models have been designed for a range of applications, such as Generative Pre-Training (GPT) (Radford et al., 2018), MOMENT (Goswami et al., 2024), or TabPFN-v2 (Hollmann et al., 2025). Although many of them have only been developed recently, foundation models have found many applications in different fields, such as engineering (Vu et al., 2025), medicine (Mazurowski et al., 2023), or education (Neumann et al., 2025).

## 3. TABULAR PRIOR-DATA FITTED NETWORK (TABPFN)

TabPFN is a powerful model specialized in tabular data (Hollmann et al., 2023; Hollmann et al. 2025), being developed using the Prior-Data Fitted Network (PFN) architecture (Müller et al., 2022; Hollmann et al., 2023; Hollmann et al. 2025). In essence, as proposed by Müller et al. (2022), PFN is a modified Transformer encoder designed to perform Bayesian inference. The architecture removes positional encodings from the Transformer to ensure permutation invariance in the dataset. In principle, PFN is trained across a large amount of artificial dataset drawn from a prior distribution $p(\mathcal{D})$, learning to approximate the posterior predictive distribution $p(y|x, D)$. Methodically, the model is trained using Prior-Data Negative Log-Likelihood (NLL) loss, which can be defined as:

$$l_\theta = E_{D \cup (x,y) \sim p(\mathcal{D})}[-log(q_\theta(y|x, D))] \qquad \text{(Equation 6)}$$

Being the loss function, Equation 6 drives the model to match its predictive distribution to the true Bayesian posterior averaged over priors (Müller et al., 2022). In this equation, $l$ denotes the loss function to be minimized, $\theta$ represents the parameters of the neural network, $E$ denotes the expectation operator, $q_\theta$ is the parameterized model, $D$ represents the dataset, $x$ denotes the input features, and $y$ corresponds to the true label associated with each input x. Additionally, $D \cup (x, y)$ is a dataset sampled from $p(\mathcal{D})$ (Müller et al., 2022).

Based on PFN architecture, Hollmann et al. (2023) leverage in-context learning, the same mechanism behind the success of large language models (LLM), to develop a fully learned tabular prediction mechanism as TabPFN. Next, Hollmann et al. (2025) improve TabPFN and introduce TabPFN-v2, which can handle larger datasets and support up to 10,000 examples for tasks like regression and classification. Instead of learning from one dataset through gradient updates as the case in conventional supervised learning, TabPFN-v2 is trained across millions of synthetic datasets, allowing it to infer new relationships directly from context without further optimization. This approach enables the model to approximate a wide range of learning algorithms, including complex Bayesian ones, through a single forward pass. The core idea is to generate massive synthetic tabular datasets using structural causal models that encode realistic dependencies, noise, categorical variables, and missing values. These synthetic datasets define the "prior" from which TabPFN-v2 learns to predict masked targets with given observed samples (Hollmann et al., 2025).

As described by Hollmann et al. (2025), structural causal models simulate real-world data-generating mechanisms by defining causal graphs with nodes (variables) and edges (functional dependencies). Random noise variables are propagated through the causal graph, applying nonlinear mappings such as neural networks, decision trees, or mathematical functions (e.g., sigmoid, ReLU, sine, modulo). Gaussian noise and post-processing steps like quantization, discretization, and warping are added to mimic real-world imperfections such as missing values, nonlinear distortions, and categorical variables. Notably, each generated dataset includes both input features and target variables some of which are masked to simulate supervised learning tasks. Altogether, this generative framework allows TabPFN-v2 to observe a rich variety of data patterns - linear, nonlinear, correlated, or noisy - and enables it to generalize across a broad spectrum of tabular problems (Hollmann et al., 2025).

Emphatically, using prepared synthetic dataset, TabPFN-v2 is pre-trained to perform masked predictions, thereby internalizing a general-purpose learning procedure (Hollmann et al., 2025).When applied to a new, unseen dataset, TabPFN receives all training and test samples at once and output predictions through in-context learning - essentially performing training and inference in one step. Its two-way attention architecture allows each cell in a table to attend both to other features in the same row and to the same feature across different rows, ensuring permutation invariance and efficient handling of tabular structure. Theoretically, similar to PFN, TabPFN-v2 approximates Bayesian posterior prediction, learning to predict $P(Y_{test}|X_{test}, Y_{train}, X_{train})$ (Hollmann et al., 2025). Overall, TabPFN-v2 transforms algorithm design into a data-driven meta-learning problem, replacing explicit programming and parameter tuning with a model that learns to learn directly from diverse, causally grounded synthetic examples, resulting in a universal and efficient tabular prediction engine.

As proven across various tasks, TabPFN-v2 has shown its promising performance across multiple test datasets (Hollmann et al., 2025). In just a few seconds, TabPFN-v2 can easily outperform even SOTA models in different tasks without requiring further adjustments (Hollmann et al., 2023; Hollmann et al., 2025). Furthermore, TabPFN-v2, as a foundation model, also allows users to perform fine-tuning for enhanced performance (Hollmann et al., 2025) and to tackle a wide range of tasks such as security (Ruiz-Villafranca et al., 2024), geotechnical (Saito et al., 2025), or medical applications (Luo et al., 2025). Moreover, Hoo et al. (2025) indicate that TabPFN-TS, a variant of TabPFN-v2, has the potential to be used for sequence data by showing its application in time-series processing. Advantageously, despite its better performance, TabPFN-TS only needs 11 million parameters, much less than billions of parameters required by other large foundation models (Tu et al., 2024; Hoo et al., 2025). These significant features highlight the robust applicability of TabPFN-v2 in various domains, which inspires us to employ TabPFN to tackle CO problems with TSP as an illustration in this paper.

## 4. METHOD

This section presents the method to solve TSP using TabPFN-v2. The process herein includes data preparation, training, solution decoding, and post-processing. It also discusses how to benchmark and evaluate the performance of the algorithm.

### 4.1 Data Preparation

In this study, to showcase the strengths of TabPFN, a single sample is generated for our adaptation and fine-tuning process. Specifically, the sample is generated by creating 500 points randomly inside a 2D unit square to represent a 500-city TSP problem.

Each point is presented by two coordinates $x \in [0,1]$ and $y \in [0,1]$. Notably, this is also a standard procedure adopted in different papers, such as that by Li et al. (2024).

Next, Google OR-Tools is selected as the TSP solver for the data generation process. Methodically, OR-Tools is chosen since it is a highly trusted software, considered to be fast, portable, and designed to solve both simple and complex CO problems using SOTA algorithms (Alves et al., 2021; Google, n.d.). Procedurally, this solver is run for 30 minutes to solve the 500-point sample, and the best-found solution is recorded for the input processing, as described in the subsequent Subsection 4.2.1.

## 4.2 Training and Solution Decoding

### 4.2.1 Input Processing

The aim of our adaptation and fine-tuning process is to learn from the optimized solution delivered by the OR-Tools solver. Hence, the input should capture the information from the optimized tour in a meaningful way. As outlined in Subsection 4.1, the input for TabPFN-v2 is based on the solution generated by OR-Tools, which produces a single optimized tour for one input sample. In principle, if the entire route of a sample is used as the input, the number of columns representing such a sample must be large enough to represent all node information and vary subject to the size of the specific TSP sample. Hence, this whole-instance approach requires separate adapted models for different TSP problem sizes. For this reason, we adopt a node-based approach, where nodes are sequentially appended to a solution one by one. This approach involves multiple node predictions within a single TSP sample, with each prediction corresponding to the model output for that specific node. To enable meaningful learning, the output must adapt to each new node input, necessitating multiple distinct outputs for each node prediction. However, since there is only one optimal solution per TSP sample, the fully optimized tour cannot be used directly as the output for training or adaptation.

Building on this, instead of predicting the entire tour at once, our node-based approach can be reframed as a sequence prediction problem, where the model predicts the potential location to visit at each step, making the output more dynamic and flexible for varying instance sizes. Accordingly, the input for the next node prediction is set to be the information of the current node position. However, it is inadequate to rely solely on single raw 2-D coordinates and edge distance data, because these features lack contextual relationships that guide real route construction. As discussed by Zhao & Wong (2025), the spatial information of the current node and its surrounding neighbours is extremely important in finding an optimal tour. Furthermore, in a TSP sample, nodes in proximity are more likely to be connected in the optimal solution (Zhao & Wong, 2025), which suggests that the information about the closest neighbours can be meaningful for the adaptation and fine-tuning process. Another possible input is the information about the previously visited nodes of the solution. However, if this information is included, multiple inference steps are required to build the TSP route since the previous node information is dependent on the existing route, which in turn can reduce the ability for parallel processing and increase inference time.

With such considerations, to enable the model to construct optimal tours, the input layer is carefully designed to encode both the current state of traversal and the local neighbourhood structure. In short, the input information includes the following:
- The current node in the path is represented by its two-dimensional coordinates. This feature provides an explicit spatial reference for the model, allowing it to evaluate where the tour is presently located and to plan the next move accordingly.
- The k-closest neighbours are represented. For each of these neighbours, the input includes the exact 2-D coordinates and the distance from each of these neighbours to the current node in the path. In addition, the selection of these neighbours is determined by the distance, and this procedure varies subject to the respective phase of the process. For clarification, during the adaptation process, the closest neighbours are selected based on their distance from the next node location. In the testing process, as there is no information on the location of the next node, the distances are calculated based on their respective distances from the current node in the path. As mentioned by Zhao & Wong (2025), such a difference in the distance calculation has minimal impact since nodes that are close to each other are often connected in the optimal solution.

In the next step, the output of the model predicts the potential location of the next node to be visited, which effectively transforms the TSP into a regression task. After all such information is calculated, it is transferred to a CSV file, which can be used to fine-tune TabPFN-v2. To illustrate the structure of the proposed input representation, the node-based embedding components are visualized in Figure 1. The current node (circle shape) serves as the focal point of prediction, while the predicted next node (triangle shape) and the actual next node (square shape) indicate the model's output and ground truth in the optimal solution, respectively. The surrounding five closest neighbour nodes (cross shape) are labelled from the closest node (Node 1) to the furthest node (Node 5). These nodes represent the local spatial context encoded in the input features of the model. Figure 1 also demonstrates that the next node to be visited in the optimal solution is Node 2, instead of the closest node (Node 1). This strategy bypasses the local greedy choice (Node 1) and selects the less favorable option (Node 2) when Node 2 exhibits higher spatial connectivity (i.e., closer to multiple nodes). By prioritizing nodes that offer greater future branching opportunities, the algorithm increases the likelihood of improved path construction in subsequent decision steps.

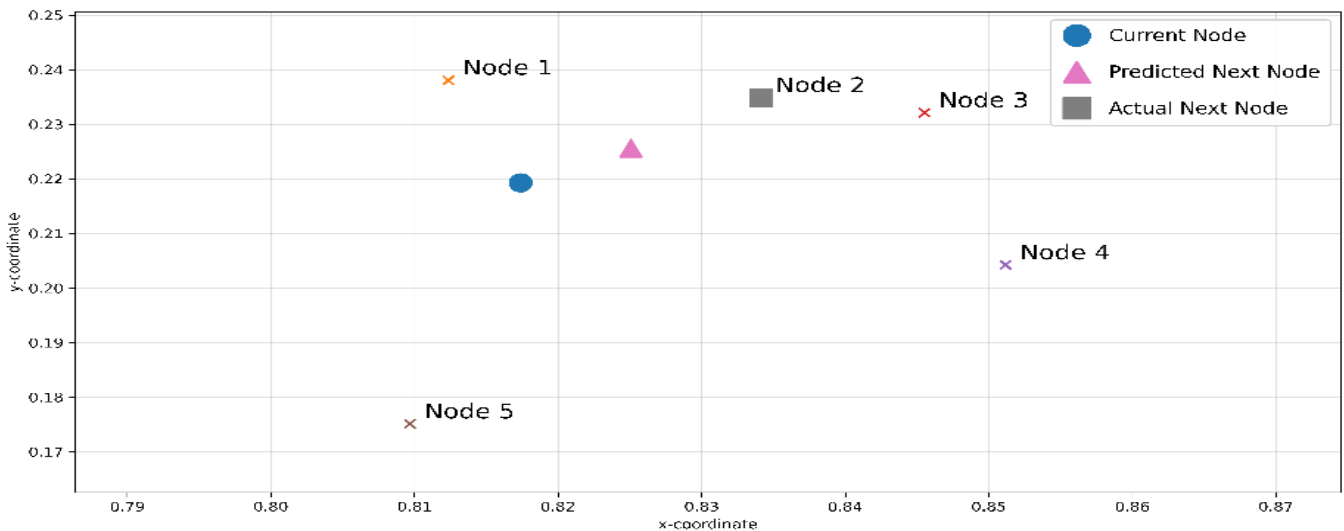

Figure 1. Visualization of Node-Based Input Embedded in a TSP Sample.

By adopting this node-based approach, the method aims at the following benefits:
- Scalability across different instance sizes: Each prediction step relies on a fixed number of nearby neighbours, preventing the input size from growing with the full graph. This significantly simplifies the model architecture and removes the need for specialized designs when testing on new instance sizes. This approach may help mitigate the generalization issue across different TSP instance sizes, a challenge discussed by Joshi et al. (2021).
- Improved emphasis on local spatial structure: By directly observing local neighbours, the model can adapt to local geometric patterns, which are often crucial in TSP, as argued by Zhao & Wong (2025).
- Reduced computational costs: Since each step only processes a small local neighbourhood, the inference for each node is lightweight, enabling more efficient computation.

*4.2.2 Model Adaptation*

The model adaptation process, or fine-tuning process, adjusts the pretrained model to the target dataset. During this process, the original fine-tuning code provided by the TabPFN authors (Hollmann et al., 2025) is used to update model parameters and align internal components for prediction, resulting in the adapted model. Then, our adapted model is created by first optimizing the number of closest neighbours used as the input and testing various configurations to determine the optimal value, using the above-mentioned adaptation process. Specifically, the number of neighbours in the k-nearest neighbours method ranges from 1 to 40, using the same 500-node sample as the input for all evaluations. Next, the best adapted TabPFN-v2 regressor model, optimized with the optimal number of closest neighbours, is saved. Notably, since TabPFN-v2 does not support multiple outputs, two distinct models are created separately to predict the x-coordinate and y-coordinate independently.

During the inference for evaluation, the same type of input created for the adaptation process (i.e., the location of each node and its k-closest neighbour) is given to the adapted model, resulting in one row for each node in each TSP evaluation sample. The adapted model then uses this same-type input to predict the potential location of the next node for each node in parallel. This prediction can subsequently be used to assess the potential of all nodes to infer the next possible stop for each node, following the method outlined in Subsection 4.2.3.

*4.2.3 Output Decoding*

As described in Subsection 4.2.1, our adapted model is designed to output the potential location $(x_i^{po}, y_i^{po})$ of the next node for the current node $i$, and each node $i$ is processed independently. However, since different test samples might have different node locations, a probabilistic mapping is required rather than rigidly matching a fixed set of locations. Hence, we need to design a decoding mechanism. Accordingly, the decoding mechanism herein is adapted from the method used by Li et al. (2024). To be more specific, the model first examines the list of nodes $\{(x_1, y_1), (x_2, y_2), (x_3, y_3), \ldots, (x_n, y_n)\}$ and assigns an unnormalized score for the current node $i$ and the other nodes using the Euclidean distance:

$$\boldsymbol{D_{ij}} = \sqrt{\left(x_i^{po} - x_j\right)^2 + (y_i^{po} - y_j)^2} \quad \text{(Equation 7)}$$

This calculation results in a matrix $D$ of size $n^2$. It should be noted that, since each node is visited exactly once, $D_{ij}$ is manually set to 0 if $i = j$. After that, any value that is infinite or equal to 0 is filtered out, and the median $\tau$ of the remaining values is calculated. The array $D$ is then normalized by dividing it by the median $\tau$, or

$$\boldsymbol{D_{normalized}} = \frac{D}{\tau} \quad \text{(Equation 8)}$$

The next step is to translate this matrix into a probability function by applying the softmax function, resulting in the probability matrix $P$. In this matrix, each entry $P(i, j)$ represents the likelihood that point $i$ is connected to point $j$. Diagonal entries are set to 0 to avoid self-loops. Notably, in this notation, the whole matrix is stacked together, resulting in a $n^2$ vector before the softmax function is applied, as shown in Equation 10.

$$\boldsymbol{P(i,j)} = \begin{cases} softmax(-D_{normalized}(i,j)) & if\ i \neq j \\ 0 & if\ i = j \end{cases} \quad \text{(Equation 9)}$$

Then, two types of candidate edges are generated:
- **Spatial k-NN edges**: Edges are generated between nodes that are spatially close.
- **Full-matrix edges**: Edges are generated between all pairs of nodes.

The final step is to create the actual TSP route. This process is completed by adding unconnected edges from the highest probability to the lowest one using the Spatial k-NN edges while ensuring that no subtour is formed, i.e., there is no Hamiltonian cycle that does not cover all nodes. If the Spatial k-NN edges cannot create a complete solution, the full-matrix edges can be used to create a feasible one.

## 4.3 Post-processing

The solution created by the decoding phase (Subsection 4.2.3) can be further improved by a simple 2-opt heuristic. Observably, 2-opt is the heuristic that is utilized in many other studies to improve initial solutions, such as those in Qiu et al. (2022), Sun & Yang (2023), or Li et al. (2024). The implementation process in this research works by picking all possible pair nodes from a complete solution and reversing the order of the route between those two nodes. The new solution is accepted if it is better than the best-known solution of a given sample, or else that new solution is discarded. The process continues until the time limit is reached, or an iteration cannot find a better solution.

## 4.4 Benchmarking and Evaluation

The benchmarking and evaluation process is based on different instance sizes. Notably, only one adapted model using a single sample for adaptation is tested for all TSP instance sizes, and any change in evaluating sizes must not require new adaptation in our research. For fair comparison, the evaluation of small instance sizes (i.e., 100 nodes or fewer) is done on two different sizes: 50 nodes (TSP-50) and 100 nodes (TSP-100), using the dataset created by Joshi et al. (2021)[5]. For larger instance sizes (i.e., 500 nodes or more), we test on two sizes of 500 nodes (TSP-500) and 1000 nodes (TSP-1000), using the test dataset created by Fu et al. (2021)[6]. Notably, the solutions in all the above-mentioned test datasets (Fu et al., 2021; Joshi et al., 2021) are created using Concorde exact solver (Applegate et al., 2003b). In addition to Concorde exact solver, we also compare our results with solutions created with LKH-3 and OR-Tools, which are two popular SOTA TSP metaheuristic solvers (Helsgaun, 2017; Google, n.d.). Noticeably, the results of LKH-3 are often comparable to those of the Concorde exact solver with the performance gap of nearly 0%, as shown in many publications (Qiu et al., 2022; Li et al., 2023; Li et al., 2024; Meng et al., 2025). Furthermore, not all of these publications present test results on all test samples with different instance sizes (TSP-50, TSP-100, TSP-500, and TSP-1000).

To ensure consistency and comparability with prior work, this paper adapts the same evaluation metrics and baseline as provided in other papers, including those by Kool et al. (2019), Li et al. (2024), and Meng et al. (2025). Additionally, the training time and number of samples used for benchmarking models from those papers and for our approach are included to facilitate our comparison. Altogether, we adopt four evaluation metrics throughout our experiments for the above-mentioned problems. They are:

1. Length: The average total distance of the optimized routes.
2. Gap: The relative gap between the best methods (i.e., the baseline) and the length of each corresponding solver. To be more specific, the gap is computed as the absolute distance difference between the baseline and the corresponding solver, divided by the route length of the baseline. Similar to the approach adopted in prior work, the Concorde exact solver (Applegate et al., 2003b) is selected as the baseline for all evaluation samples in this research.

---

[5] Dataset available at https://github.com/chaitjo/learning-tsp
[6] Dataset available at https://github.com/SaneLYX/TSP_Att-GCRN-MCTS

3. Training or adaptation time required for each model: Notably, this metric is only applicable to ML-based solvers, including TabPFN-v2.
4. Data usage: The number of TSP samples used for training for each method and each instance size. This metric is only applicable to ML-based solvers, including TabPFN-v2.

Noticeably, the inference time is not reported as a metric, as all ML-based solvers produce solutions within a few seconds per sample using simple greedy decoding. Thus, the difference in inference time between methods is considered negligible.

It should be emphasized that, even when different papers can share the same evaluation metrics, their test results might vary subject to different test attempts due to:

1. Differences in generated samples: Since samples are often generated randomly, the actual optimal length for each sample might be different from that of the others. For this reason, this paper uses the gaps of each solver and the baseline for each instance as the main metrics for performance comparison.
2. Differences in hardware: Different research might use different hardware resources, especially the GPU, resulting in varied performance.
3. Differences in test size: Different papers might use different sizes of test instances, resulting in different test times.
4. Effects of parallelism: Most of the available public codes for published papers use parallel processing to improve inference time and make full use of available hardware. This can improve performance, especially solving time, when multiple GPUs are available.

Hence, to make our comparisons consistent and fair, the results of our method are calculated based on the average of 30 sample cases. This is because the central limit theorem is applicable when the sample size is 30 or more, as suggested by Chang & Lee (2017). Furthermore, to demonstrate the performance of a SOTA ML solver with limited training resources, we retrain the Fast-T2T model as proposed by Li et al. (2024) using 50,000 samples for two different instance sizes: 50 nodes (TSP-50) and 100 nodes (TSP-100), resulting in a total of 100,000 training samples used across these two instance sizes. The retrained models achieved for these two instance sizes are referred to in this paper as Fast-T2T 50-node and 100-node model, respectively. Rationally, this model by Li et al. (2024) is selected for retraining as it is one of the most recent SOTA solvers for TSP with the source code available online[7]. Additionally, to ensure a fairer comparison, we present the test results of the Fast-T2T models using the pretrained models provided by the original authors (Li et al., 2024), which we refer to as the original Fast-T2T model in this paper.

## 5. EXPERIMENT RESULTS AND DISCUSSION

Presenting our experimental results, this section starts with small TSP instance sizes (i.e., TSP-50 and TSP-100), which are experimented to identify the effect of the number of closest neighbours on the quality of output solutions (Subsection 5.1). Accordingly, Subsection 5.2 presents the evaluation results on all four instance sizes (TSP-50, TSP-100, TSP-500, and TSP-1000). Finally, Subsection 5.3 discusses the results presented in Subsections 5.1 and 5.2. All experiment results are performed on a system equipped with Intel Xeon Gold 5218 CPU and Nvidia A40 GPU.

---

[7] The code associated with the study by Li et al. (2024) is available at https://github.com/Thinklab-SJTU/Fast-T2T.

## 5.1 Effects of the Number of Closest Neighbours

Based on the input processing and adaptation method presented in Subsection 4.2, this subsection explores whether varying the number of closest neighbours can influence the model performance. Notably, this part tests the effect only on small instance sizes (i.e., TSP-50 and TSP-100). Purposefully, we do not evaluate the effect on larger instance sizes (e.g., TSP-500 and TSP-1000) to reduce possible risks of bias caused by size-specific adjustments during evaluation, which contrasts with approaches like Fast-T2T (Li et al., 2024) that require specialized models for each instance size. As specified in Subsection 4.2.2, the number of closest neighbours varies from 1 to 40. Notably, to ensure statistical reliability, results are evaluated by calculating the average gap across 30 instances randomly sampled from the dataset by Joshi et al. (2021).

As illustrated in Figure 2, the performance varies significantly with the number of closest neighbours, underscoring the necessity of optimizing this hyperparameter. When the number of closest neighbours is too small, the model does not have enough information to learn, resulting in poor performance. In contrast, when the number of closest neighbours is significantly large, the model might be affected negatively by less useful information, resulting in downgraded results. Observably from Figure 2, when the number of closest neighbours $k$ is between four and eight, the performance is relatively similar. Notably, this optimal number $k$ may differ for specific cases. As shown in Figure 2, the best performance is achieved when $k$ is five. In other words, the optimal performance might be attainable when the number of closest neighbours is set to five. Accordingly, all subsequent experiments in this study adopt five closest neighbours to ensure consistency and optimality.

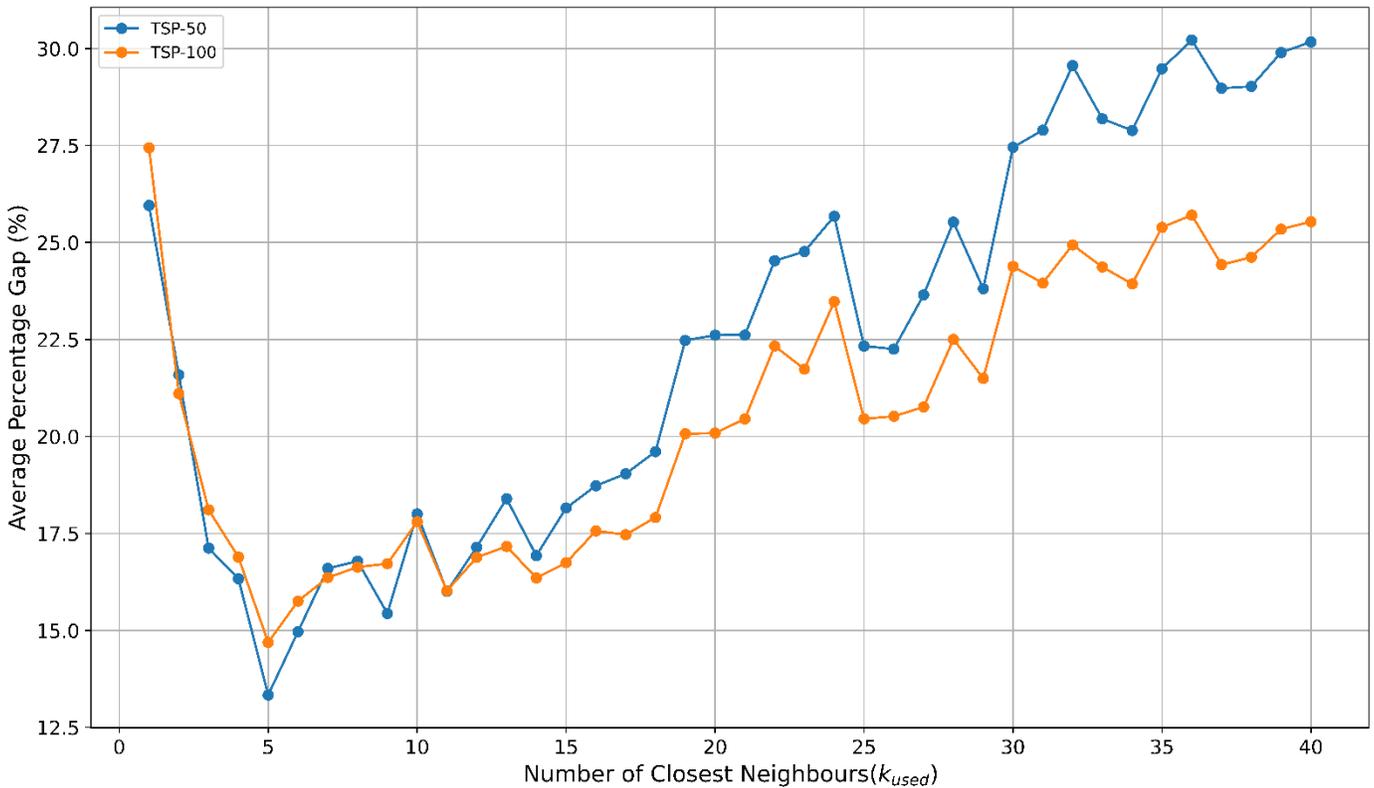

Figure 2. Effect of Closest Neighbour Size on Average Percentage Gap.

## 5.2 Evaluation Results

Using the four evaluation metrics as stated in Subsection 4.4, Table 1 presents the comprehensive evaluation of Concorde exact solver, heuristic solvers, SOTA ML-based solvers, and TabPFN-v2 for smaller instance sizes (TSP-50 and TSP-100). Importantly, the gaps presented in this subsection are compared against Concorde exact solver as the baseline. In general, Concorde and other heuristics solvers such as OR-Tools or LKH-3 can achieve relatively strong performance, and their performance differences are relatively insignificant (2.17% gap or less). Emphatically, other learning methods, i.e., DIFUSCO (Sun & Yang, 2023), T2T (Li et al., 2023), Fast-T2T (Li et al., 2024), Eformer (Meng et al., 2025), TabPFN-v2 (Hollmann et al., 2025), demonstrate the potential to achieve competitive performance in terms of solution quality, especially with the help of simple 2-opt post-processing, as evidenced by the gaps of approximately 3% or less between TabPFN-v2, Fast-T2T, and Concorde.

Overall, for these smaller instance sizes (TSP-50 and TSP-100), TabPFN-v2's performance is comparable to other ML-based SOTA solvers whether it is with or without the help of 2-opt post-processing. Notably, even without 2-opt post-processing, TabPFN-v2 outperforms DIFUSCO and T2T in a number of cases. For example, for TSP-100, DIFUSCO with $T_s = 1$ achieves 20.2% gap, seeing TabPFN-v2 outperforming at 15.36%. The performance difference between TabPFN-v2 (15.36%) and retrained Fast-T2T (13.96%) is around 1%. TabPFN-v2 exhibits lower performance compared to solvers like AM (Kool et al., 2019), original Fast-T2T (Li et al., 2024), and Eformer (Meng et al., 2025) on these smaller instance sizes. However, this performance gap can be narrowed down significantly through the application of simple 2-opt post-processing. From Table 1, TabPFN-v2 with 2-opt achieves performance similar to that of other ML-based models combined with 2-opt, as evidenced by gap differences of less than 1% between TabPFN-v2 and retrained Fast-T2T, and approximately 2% compared to the Concorde exact solver baseline.

With such performance, TabPFN-v2 shows distinct advantages in terms of training time and data requirements. To be more specific, as shown in Table 1, our findings demonstrate that ML-based models, except for TabPFN-v2, require significant amount of data (from 25,600 samples to more than 100 million samples), and long training time (from 6 hours up to more than 20 days). From our observation, when training resources are limited (i.e., fewer samples and less training time), the performance of current SOTA ML models deteriorate significantly. This is evidenced by performance gaps ranging from 5% and 10% between our retrained Fast-T2T models, and the version provided by the original authors (Li et al., 2024) on both TSP-50 and TSP-100. In contrast, TabPFN-v2 adapts to the TSP in just two minutes using a single sample, and it even outperforms SOTA ML models in some cases, as evidenced in the comparison between TabPFN-v2, T2T, and DIFUSCO in our experiments.

With the same four metrics provided in Subsection 4.4, Table 2 shows the evaluation results of TabPFN-v2 on larger instance sizes (TSP-500 and TSP-1000), revealing a trend similar to that of Table 1. Numerically, Concorde and heuristic solvers still perform the best, as performance gaps between Concorde, LKH-3, and OR-Tools, are relatively small (4% or less). Notably, Concorde results are slightly worse than LKH-3. Practically, it can be due to precision errors during the calculation process which has also been experienced in previous literature, such as TSP-10000 results in research by Zhou et al. (2025). However, this error is relatively small, as shown by the gap of almost 0% between these two solvers. However, ML-based methods produce significantly varied results. Indeed, performance gaps for these larger instance sizes for AM (Kool et al., 2019) and Eformer (Meng et al., 2025) increase significantly, being between 20-30% for TSP-500 and over 30% on TSP-1000 compared to Concorde exact solver (Applegate et al., 2003b).

**Table 1. Results with Greedy Decoding on TSP-50 and TSP-100. SL: Supervised Learning, RL: Reinforcement Learning, G: Greedy Decoding, $T_s$: Number of Sampling Steps, $T_g$: Number of Gradient Search Steps, x1: Greedy Inference. * denotes results that are quoted from previous works**

| Method/Model Name | Type | TSP-50 | | | | TSP-100 | | | |
|---|---|---|---|---|---|---|---|---|---|
| | | Length | Gap (%) | Training or Adaptation Time | Data Used (Samples) | Length | Gap (%) | Training or Adaptation Time | Data Used (Samples) |
| Concorde (baseline) (Joshi et al., 2021) | Exact | 5.65 | 0.00 | N/A | N/A | 7.70 | 0.00 | N/A | N/A |
| LKH-3 (Helsgaun, 2017) | Heuristics | 5.65 | 0.00 | N/A | N/A | 7.70 | 0.00 | N/A | N/A |
| OR-Tools (Google, n.d.) | Heuristics | 5.66 | 0.03 | N/A | N/A | 7.87 | 2.17 | N/A | N/A |
| AM* (Kool et al., 2019) | RL+G | 5.80 | 1.76 | 27hr 13m | 128,000,000 | 8.12 | 4.53 | 45h 50m | 64,000,000 |
| DIFUSCO* (Sun & Yang, 2023; Li et al., 2024) $T_s = 1$ | SL+G | 6.42 | 12.84 | N/A | 1,502,000 | 9.32 | 20.20 | N/A | 1,502,000 |
| DIFUSCO* (Sun & Yang, 2023; Li et al., 2024) $T_s = 100$ | SL+G | 5.71 | 0.41 | N/A | 1,502,000 | 7.84 | 1.16 | N/A | 1,502,000 |
| T2T* (Li et al., 2023; Li et al., 2024) $T_s = 1, T_g = 1$ | SL+G | 6.15 | 8.15 | 56h 24m | 1,502,000 | 9.00 | 16.09 | 206h 20m | 1,502,000 |
| T2T* (Li et al., 2023; Li et al., 2024) $T_s = 50, T_g = 30$ | SL+G | 5.69 | 0.03 | 56h 24m | 1,502,000 | 7.76 | 0.11 | 206h 20m | 1,502,000 |
| Fast-T2T (Li et al., 2024) 50-node model | SL+G | 5.91 | 4.52 | 6h 40m | 50,000 | 8.78 | 13.96 | 10h 25m | 50,000 |
| Fast-T2T (Li et al., 2024) 100-node model | SL+G | 5.94 | 5.04 | 6h 40m | 50,000 | 8.18 | 6.25 | 10h 25m | 50,000 |
| Fast-T2T (Li et al., 2024) Original model | SL+G | 5.69 | 0.31 | 112h 45m | 1,502,000 | 7.76 | 1.32 | 488h 12m | 1,502,000 |
| Fast-T2T with 2-opt (Li et al., 2024) 50-node model | SL+G+2-opt | 5.70 | 1.03 | 6h 40m | 50,000 | 7.95 | 3.1 | 10h 25m | 50,000 |
| Fast-T2T with 2-opt (Li et al., 2024) 100-node model | SL+G+2-opt | 5.71 | 1.19 | 6h 40m | 50,000 | 7.83 | 1.62 | 10h 25m | 50,000 |
| Eformer (x1)* (Meng et al., 2025) | SL+G | 5.70 | 0.13 | N/A | 51,000,000 | 7.79 | 0.32 | N/A | 101,000,000 |
| TabPFN | SL+G | 6.40 | 13.27 | 2m | 1 | 8.88 | 15.36 | 2m | 1 |
| TabPFN with 2-opt | SL+G+2-opt | 5.77 | 2.07 | 2m | 1 | 7.90 | 2.59 | 2m | 1 |

**Table 2. Results with Greedy Decoding on TSP-500 and TSP-1000. SL: Supervised Learning, RL: Reinforcement Learning, G: Greedy Decoding, $T_s$: Number of Sampling Steps, $T_g$: Number of Gradient Search Steps, x1: Greedy Inference. * denotes results that are quoted from previous works**

| Method/Model Name | Type | TSP-500 | | | | TSP-1000 | | | |
|---|---|---|---|---|---|---|---|---|---|
| | | Length | Gap (%) | Training or Adaptation Time | Data Used (Samples) | Length | Gap (%) | Training or Adaptation Time | Data Used (Samples) |
| Concorde (baseline) (Fu et al., 2021) | Exact | 16.58 | 0.00 | N/A | N/A | 23.23 | 0.00 | N/A | N/A |
| LKH-3 (Helsgaun, 2017) | Heuristics | 16.55 | 0.00 | N/A | N/A | 23.13 | 0.00 | N/A | N/A |
| OR-Tools (Google, n.d.) | Heuristics | 17.08 | 3.02 | N/A | N/A | 24.17 | 3.72 | N/A | N/A |
| AM * (Kool et al., 2019; Li et al., 2024) | RL+G | 20.02 | 20.99 | N/A | N/A | 31.15 | 34.75 | N/A | N/A |
| DIFUSCO * (Sun & Yang, 2023; Li et al., 2024) $T_s = 100$ | SL+G | 18.17 | 9.82 | N/A | 128,000 | 25.74 | 11.36 | N/A | 64,000 |
| Fast-T2T (Li et al., 2024) Original model | SL+G | 17.55 | 6.10 | 142h 17m | 128,000 | 24.63 | 6.53 | 324h 43m | 64,000 |
| Eformer (x1)* (Meng et al., 2025) | SL+G | 21.28 | 28.81 | N/A | N/A | N/A | N/A | N/A | N/A |
| TabPFN | SL+G | 19.84 | 19.66 | 2m | 1 | 28.55 | 22.90 | 2m | 1 |
| TabPFN with 2-opt | SL+G+2-opt | 17.12 | 3.27 | 2m | 1 | 24.36 | 5.31 | 2m | 1 |

At this stage, TabPFN-v2 is also observed to be prominent for two distinguishable reasons, as evidenced by the performance gaps of less than 20% on TSP-500 and slightly over 20% on TSP-1000. Firstly, while TabPFN-v2 uses a single training sample across all four instance sizes, other SOTA models still rely on a relatively large number of training samples when such information is available. However, the number of training samples used by these SOTA models is observably smaller than that for TSP-50 and TSP-100 as shown in Table 1, likely due to high costs of data collection. For example, DIFUSCO (Sun & Yang, 2023) and Fast-T2T (Li et al., 2024) use 128,000 samples and 64,000 samples for TSP-500 and TSP-1000 training, respectively, compared to over 1.5 million on TSP-50 and TSP-100. The reduced amount of data might be one of the explanations for the degraded performance of other SOTA supervised learning models, such as AM (Kool et al., 2019) and Eformer (Meng et al., 2025), as evidenced by the performance drop from less than 5% on smaller sizes to over 20% on larger ones. Notably, despite the reduction in the number of training samples, the training time remains relatively high, as demonstrated by the nearly 15-day training duration for original Fast-T2T (Li et al., 2024) on TSP-1000. Secondly, using only a single sample for adaptation, TabPFN-v2, with the gap of 19.66%, keeps performing better than the others, especially compared to AM (20.99%) and Eformer (28.81%). On TSP-1000, TabPFN-v2 (22.90%) outperforms AM (34.75%) by more than 10 percentage points, while Meng et al. (2025) does not report the results of Eformer for this instance size. Noticeably, all of these gap analyses are reported using Concorde exact solver (Applegate et al., 2003b) as the baseline.

### 5.3 Analysis and Discussion

The results presented in Tables 1 and 2 highlight the consistent success of our experimental application of TabPFN-v2. Fundamentally, it is possible to say that TabPFN-v2, which is a foundation model originally designed for tabular data, can be successfully adapted to solve CO problems, particularly TSP in varying instance sizes as the key target of our research, especially with time and data efficiency. To do so, our emphasis is specifically placed on a node-based encoding approach to better scalability and generalization, as well as on an efficient node-predicting adaptation strategy to extract local spatial information. To the best of our knowledge, our application is the first of its kind in harnessing the powerful and advantageous features of TabPFN-v2 to solve CO problems in general and TSP in particular.

Throughout our experiment process, although it is not originally designed for sequential, CO, or graph-based inputs, TabPFN-v2 achieves comparable or better performance to SOTA learning-based solvers across all four instance sizes (TSP-50, TSP-100, TSP-500, and TSP-1000), resulting in multiple implications. Firstly, as shown in Tables 1 and 2, the performance is achieved without extensive training, relying on a single sample across all different instance sizes and a two-minute adaptation phase. Possible underlying reasons can include the capacity of TabPFN-v2 to guide the algorithm away from the seemingly best immediate step by leveraging strengths of in-context learning if that greedy move leads to a suboptimal solution, as illustrated in Figure 1. Inferably, this beneficial capacity may help address the generalization issue across different TSP instance sizes, a phenomenon highlighted by Joshi et al. (2021). Secondly, the above-mentioned performance of TabPFN-v2 highlights the value of rapid adaptation and training efficiency as one of the core advantages inherent in our method, which are crucial in practical settings where training data is scarce, expensive, or time-consuming to obtain. Indeed, it might be impossible sometimes to collect millions of training samples required for these supervised learning solvers in real-world scenarios. For example, in the research by Cappart et al. (2018), the dataset that the authors use has only one sample per day. Thirdly, TabPFN-v2 can be very useful when it comes to resource requirements observed in most of the potentially competing ML-based models. Contrastively, as briefly discussed in Section 1 and Subsection 2.1.3, these models often require substantial resources, including tens of thousands to millions of training samples and long training sessions ranging from several hours to multiple days. Fourthly, conventional ML-based models should be specific for given instance sizes, or else risk poor performance will result. Even if the model is retrained for each instance size, the same model size can further limit the effectiveness of sample encoding, especially for large-size problems. This phenomenon might explain the significant performance degradation as the instance size increases, as shown in the case of AM (Kool et al., 2019), and Eformer (Meng et al., 2025). In contrast, TabPFN-v2, as a foundation model, is pre-trained on multiple datasets to learn patterns of tabular data, which helps it to adapt to new problems faster and utilize domain-specific data more efficiently while maintaining comparable performance. Such performance differences can particularly be seen in the results discussed in Subsection 5.2.

Methodically, another significant contribution derived from our proposed application of TabPFN-v2 is the use of a node-based encoding approach, as opposed to whole-instance encoding methods. In essence, this encoding approach can leverage the above-mentioned in-context learning capacity of TabPFN-v2 to improve the generalization ability of the model, especially when scaling from small to large TSP samples. Using the Concorde exact solver (Applegate et al., 2003b) to be the baseline as established in Subsection 4.4, our results confirm that TabPFN-v2 can maintain relatively strong performance for varied instance sizes although we keep using a single adapted model. This is evidenced by smaller gaps of less than 20% on TSP-500 and slightly over 20% on TSP-1000 versus the baseline. In contrast, whole-instance encoding models such as those proposed by Meng et al. (2025) often experience performance degradation when the same whole-instance embedding architecture is applied to different instance sizes. This limitation is evident in our experiments on TSP-500 and TSP-1000, where methods such as Eformer (Meng et al., 2025) exhibit performance gaps of approximately 30%. This underscores the flexibility and scalability of our node-based approach, making it more suitable for real-world applications where instance sizes may vary widely.

Furthermore, our proposed method opts for adapting TabPFN-v2 to predict the next node in a tour, which also proves to be effective across both small-sized and large-sized TSP problems. Even without 2-opt post-processing, our adapted model can be equally good or even outperform several recent ML-based approaches. For example, applying the above-mentioned baseline, TabPFN-v2 demonstrates a performance gap on TSP-500 that is almost 10% lower than the results reported by the authors of Eformer (Meng et al., 2025). When 2-opt is applied, the overall solution quality improves, and the performance gap between TabPFN-v2 and other ML-based models becomes negligible, illustrated by a gap difference of less than 1% TabPFN-v2 and Fast-T2T.

## 6. CONCLUSION

This paper explores and proposes an application of TabPFN-v2, a supervised tabular foundation model, to Combinatorial Optimization (CO) problems, specifically the Traveling Salesman Problem (TSP) as the key target of our research. To the best of the authors' knowledge, this can be the first application of its kind in using TabPFN-v2 to deal with CO problems in general and TSP in particular. The core of our method and experiments is the use of a novel node-based encoding approach to avoid expressiveness issues, the lack of local spatial structure, and to enhance the scalability in combination with the node-predicting adaptation strategy to predict the location of the next node in constructing the entire TSP route. Thereby, we demonstrate that TabPFN-v2 can be effectively applied to routing-related tasks such as those in engineering, biology, or economics (Johnson & Liu, 2006; Crişan et al., 2020; Singh et al., 2022), achieving performance competitive with, or even better, than existing state-of-the-art (SOTA) Machine Learning (ML)-based solvers.

Three key strengths should be highlighted in our proposed application of TabPFN-v2 herein. First, it requires minimal training, being able to get adapted to TSP within minutes using only one single sample and to complete the inference process during evaluation in seconds, which leads to rapid adaptation and enhanced data efficiency. This stands in contrast to most learning-based methods, including those using training exact and heuristic algorithms and ML, which are often claimed to be considerably time and data-intensive as discussed particularly in Section 1, Subsections 2.1.2, 2.1.3, 5.2, and 5.3. Second, the node-based representation allows the model to achieve better generalization across varying instance sizes and to reduce the performance degradation commonly seen in whole-instance encoding approaches. Third, the proposed node-predicting adaptation strategy allows TabPFN-v2 to efficiently generate TSP tours in a sequential and step-wise manner. Overall, these strengths eliminate the need for extensive retraining or iterative optimization, significantly reducing computational and data requirements, while maintaining competitive solution quality and enabling rapid inference suitable for real-time applications.

Our experiments across both small and large-scale TSP samples confirm that our approach using TabPFN-v2 can deliver strong solution quality, with or without 2-opt post-processing, and achieves performance comparable to other models when simple 2-opt refinement is applied. These findings open up a promising direction for extending the use of powerful, pre-trained foundation models like TabPFN-v2 to broader classes of structured and combinatorial problems, particularly where training resources are limited, or rapid deployment is essential.

In this line, to harness the robustness of TabPFN-v2, future research could investigate alternative methods for integrating richer information capturing the state of visited and unvisited nodes, structural representations of partial tours or previously visited paths, and statistical descriptors of node location and spatial density distributions directly into TabPFN-v2 to enhance performance. Moreover, extending this framework to other graph-based combinatorial problems beyond TSP, such as vehicle routing problem (VRP) and its variants, represents a natural and valuable extension.

**Acknowledgment**

The research grant from MITACS (project number: IT40086) is gratefully acknowledged.

**DECLARATION OF COMPETING INTEREST**

The authors declare that they have no known competing financial interests or personal relationships that could have appeared to influence the work reported in this paper.
**REFERENCES**

Alanzi, E., & Menai, M. E. B. (2025). Solving the Traveling Salesman Problem with Machine Learning: A Review of Recent Advances and Challenges. *Artificial Intelligence Review*, *58*(9). https://doi.org/10.1007/s10462-025-11267-x

Alves, F., Pacheco, F., Rocha, A. M. A. C., Pereira, A. I., & Leitão, P. (2021). Solving a Logistics System for Vehicle Routing Problem Using an Open-Source Tool. In O. Gervasi, B. Murgante, S. Misra, C. Garau, I. Blečić, D. Taniar, B. O. Apduhan, A. M. A. C. Rocha, E. Tarantino, & C. M. Torre (Eds.), *Computational Science and Its Applications -- ICCSA 2021* (Vol. 12953, pp. 397–412). Springer, Cham. https://doi.org/10.1007/978-3-030-86976-2_27

Applegate, D., Bixby, R., Chvátal, V., & Cook, W. (2003a). Implementing the Dantzig-Fulkerson-Johnson Algorithm for Large Traveling Salesman Problems. *Mathematical Programming*, *97*(1), 91–153. https://doi.org/10.1007/s10107-003-0440-4

Applegate, D., Bixby, R., Chvátal, V., & Cook, W. (2003b). *Concorde TSP Solver*. https://www.math.uwaterloo.ca/tsp/concorde/

Balas, E., & Toth, P. (1983). *Branch and Bound Methods for the Traveling Salesman Problem*. https://doi.org/10.21236/ADA126957

Bektas, T. (2006). The Multiple Traveling Salesman Problem: An Overview of Formulations and Solution Procedures. *Omega*, *34*(3), 209–219. https://doi.org/10.1016/j.omega.2004.10.004

Bommasani, R., Hudson, D. A., Adeli, E., Altman, R., Arora, S., von Arx, S., Bernstein, M. S., Bohg, J., Bosselut, A., Brunskill, E., Brynjolfsson, E., Buch, S., Card, D., Castellon, R., Chatterji, N., Chen, A., Creel, K., Davis, J. Q., Demszky, D., … Liang, P. (2022). *On the Opportunities and Risks of Foundation Models*. http://arxiv.org/abs/2108.07258

Brown, T. B., Mann, B., Ryder, N., Subbiah, M., Kaplan, J., Dhariwal, P., Neelakantan, A., Shyam, P., Sastry, G., Askell, A., Agarwal, S., Herbert-Voss, A., Krueger, G., Henighan, T., Child, R., Ramesh, A., Ziegler, D., Wu, J., Winter, C., … Amodei, D. (2020). Language Models Are Few-Shot Learners. In H. Larochelle, M. Ranzato, R. Hadsell, M. F. Balcan, & H. Lin (Eds.), *Advances in Neural Information Processing Systems* (Vol. 33, pp. 1877–1901). Curran Associates, Inc. https://proceedings.neurips.cc/paper_files/paper/2020/file/1457c0d6bfcb4967418bfb8ac142f64a-Paper.pdf

Carpaneto, G., Fischetti, M., & Toth, P. (1989). New Lower Bounds For The Symmetric Travelling Salesman Problem. *Mathematical Programming*, *45*(1–3), 233–254. https://doi.org/10.1007/BF01589105

Cappart, Q., Thomas, C., Schaus, P., & Rousseau, L. M. (2018). A Constraint Programming Approach For Solving Patient Transportation Problems. In J. Hooker (Ed.), *Lecture Notes in Computer Science: Vol. 11008 LNCS* (pp. 490–506). Springer International Publishing. https://doi.org/10.1007/978-3-319-98334-9_32

Cappart, Q., Chételat, D., Khalil, E. B., Lodi, A., Morris, C., & Veličković, P. (2023). Combinatorial Optimization And Reasoning With Graph Neural Networks. *Journal of Machine Learning Research*, *24*(1), 130. https://dl.acm.org/doi/pdf/10.5555/3648699.3648829
19

Chang, H. J., & Lee, M. C. (2017). Applying Computer Simulation to Analyze the Normal Approximation of Binomial Distribution. *Journal of Computers (Taiwan)*, *28*(5), 116–131. https://doi.org/10.3966/199115992017102805011

Crişan, G. C., Pintea, C. M., Pop, P. C., & Matei, O. (2020). Economical Connections Between Several European Countries Based On TSP Data. *Logic Journal of the IGPL*, *28*(1), 33–44. https://doi.org/10.1093/jigpal/jzz069

Das, A., Kong, W., Sen, R., & Zhou, Y. (2024). A Decoder-only Foundation Model for Time-series Forecasting. *Proceedings of the 41st International Conference on Machine Learning*. https://openreview.net/forum?id=jn2iTJas6h

Frieze, A., Karp, R. M., & Reed, B. (1995). When Is the Assignment Bound Tight for the Asymmetric Traveling-Salesman Problem? *SIAM Journal on Computing*, *24*(3), 484–493. https://doi.org/10.1137/S0097539792235384

Fu, Z.-H., Qiu, K.-B., & Zha, H. (2021). Generalize a Small Pre-trained Model to Arbitrarily Large TSP Instances. *Proceedings of the AAAI Conference on Artificial Intelligence*, *35*(8), 7474–7482. https://doi.org/10.1609/aaai.v35i8.16916

Google. (n.d.). *Introduction to Optimization*. Retrieved September 1, 2025, from https://developers.google.com/optimization/introduction

Goswami, M., Szafer, K., Choudhry, A., Cai, Y., Li, S., & Dubrawski, A. (2024). MOMENT: A Family of Open Time-series Foundation Models. In R. Salakhutdinov, Z. Kolter, K. Heller, A. Weller, N. Oliver, J. Scarlett, & F. Berkenkamp (Eds.), *Proceedings of the 41st International Conference on Machine Learning*. JMLR.org. https://proceedings.mlr.press/v235/goswami24a.html

Hajkowicz, S., Sanderson, C., Karimi, S., Bratanova, A., & Naughtin, C. (2023). Artificial Intelligence Adoption In The Physical Sciences, Natural Sciences, Life Sciences, Social Sciences And The Arts And Humanities: A bibliometric analysis of research publications from 1960-2021. *Technology in Society*, *74*. https://doi.org/10.1016/j.techsoc.2023.102260

Held, M., & Karp, R. M. (1962). A Dynamic Programming Approach to Sequencing Problems. *Journal of the Society for Industrial and Applied Mathematics*, *10*(1), 196–210. https://doi.org/10.1137/0110015

Helsgaun, K. (2017). *An Extension of the Lin-Kernighan-Helsgaun TSP Solver for Constrained Traveling Salesman and Vehicle Routing Problems* (Issue 12). https://webhotel4.ruc.dk/~keld/research/LKH-3/LKH-3_REPORT.pdf

Hollmann, N., Müller, S., Eggensperger, K., & Hutter, F. (2023). TabPFN: A Transformer that Solves Small Tabular Classification Problems in a Second. *International Conference on Learning Representations (ICLR)*. https://openreview.net/forum?id=cp5PvcI6w8_

Hollmann, N., Müller, S., Purucker, L., Krishnakumar, A., Körfer, M., Hoo, S. B., Schirrmeister, R. T., & Hutter, F. (2025). Accurate Predictions On Small Data With A Tabular Foundation Model. *Nature*, *637*(8045), 319–326. https://doi.org/10.1038/s41586-024-08328-6

Hoo, S. B., Müller, S., Salinas, D., & Hutter, F. (2025). *From Tables to Time: How TabPFN-v2 Outperforms Specialized Time Series Forecasting Models*. http://arxiv.org/abs/2501.02945

Jaillet, P., & Lu, X. (2011). Online Traveling Salesman Problems With Service Flexibility. *Networks*, *58*(2), 137–146. https://doi.org/10.1002/net.20454

Johnson, O., & Liu, J. (2006). A Traveling Salesman Approach For Predicting Protein Functions. *Source Code for Biology and Medicine*, *1*. https://doi.org/10.1186/1751-0473-1-3

Joshi, C. K., Cappart, Q., Rousseau, L. M., & Laurent, T. (2021). Learning TSP Requires Rethinking Generalization. In L. D. Michel (Ed.), *Leibniz International Proceedings in Informatics, LIPIcs* (Vol. 210, pp. 33:1-33:21). Schloss Dagstuhl- Leibniz-Zentrum fur Informatik GmbH, Dagstuhl Publishing. https://doi.org/10.4230/LIPIcs.CP.2021.33

Joshi, C. K., Laurent, T., & Bresson, X. (2019). *An Efficient Graph Convolutional Network Technique for the Travelling Salesman Problem*. http://arxiv.org/abs/1906.01227

Kool, W., van Hoof, H., & Welling, M. (2019). Attention, Learn to Solve Routing Problems! *International Conference on Learning Representations (ICLR)*. https://openreview.net/forum?id=ByxBFsRqYm

Kwon, Y.-D., Choo, J., Kim, B., Yoon, I., Gwon, Y., & Min, S. (2020). POMO: Policy Optimization with Multiple Optima for Reinforcement Learning. In H. Larochelle, M. Ranzato, R. Hadsell, M. F. Balcan, & H. Lin (Eds.), *Advances in Neural Information Processing Systems* (Vol. 33, pp. 21188–21198). Curran Associates, Inc. https://proceedings.neurips.cc/paper_files/paper/2020/file/f231f2107df69eab0a3862d50018a9b2-Paper.pdf

Laporte, G. (1992). The traveling salesman problem: An Overview Of Exact And Approximate Algorithms. *European Journal of Operational Research*, *59*(2), 231–247. https://doi.org/10.1016/0377-2217(92)90138-Y

Larrañaga, P., Kuijpers, C. M. H., Murga, R. H., Inza, I., & Dizdarevic, S. (1999). Genetic Algorithms for the Travelling Salesman Problem: A Review of Representations and Operators. *Artificial Intelligence Review*, *13*(2), 129–170. https://doi.org/10.1023/A:1006529012972

Li, Y., Guo, J., Wang, R., & Yan, J. (2023). T2T: From Distribution Learning in Training to Gradient Search in Testing for Combinatorial Optimization. In A. Oh, T. Naumann, A. Globerson, K. Saenko, M. Hardt, & S. Levine (Eds.), *Advances in Neural Information Processing Systems* (Vol. 36, pp. 50020–50040). Curran Associates, Inc. https://proceedings.neurips.cc/paper_files/paper/2023/hash/9c93b3cd3bc60c0fe7b0c2d74a2da966-Abstract-Conference.html

Li, Y., Guo, J., Wang, R., Zha, H., & Yan, J. (2024). Fast T2T: Optimization Consistency Speeds Up Diffusion-Based Training-to-Testing Solving for Combinatorial Optimization. In A. Globerson, L. Mackey, D. Belgrave, A. Fan, U. Paquet, J. Tomczak, & C. Zhang (Eds.), *Advances in Neural Information Processing Systems* (Vol. 37, pp. 30179–30206). Curran Associates, Inc. https://proceedings.neurips.cc/paper_files/paper/2024/file/352b13f01566ae34affacc60e98c16af-Paper-Conference.pdf

Lu, Y., Hao, J. K., & Wu, Q. (2022). Solving The Clustered Traveling Salesman Problem Via Traveling Salesman Problem Methods. *PeerJ Computer Science*, *8*. https://doi.org/10.7717/PEERJ-CS.972

Luo, J., Yuan, Y., & Xu, S. (2025). *TIME: TabPFN-Integrated Multimodal Engine for Robust Tabular-Image Learning*. http://arxiv.org/abs/2506.00813

Mazurowski, M. A., Dong, H., Gu, H., Yang, J., Konz, N., & Zhang, Y. (2023). Segment Anything Model For Medical Image Analysis: An Experimental Study. *Medical Image Analysis*, *89*. https://doi.org/10.1016/j.media.2023.102918

Meng, D., Cao, Z., Wu, Y., Hou, Y., Ge, H., & Zhang, Q. (2025). EFormer: An Effective Edge-based Transformer for Vehicle Routing Problems. In J. Kwok (Ed.), *Proceedings of the Thirty-Fourth International Joint Conference on Artificial Intelligence, IJCAI-25* (pp. 8582–8590). https://doi.org/10.24963/ijcai.2025/954

Müller, S., Hollmann, N., Arango, S. P., Grabocka, J., & Hutter, F. (2022). Transformers Can Do Bayesian Inference. *International Conference on Learning Representations (ICLR)*. https://openreview.net/forum?id=KSugKcbNf9

Neumann, A. T., Yin, Y., Sowe, S., Decker, S., & Jarke, M. (2025). An LLM-Driven Chatbot in Higher Education for Databases and Information Systems. *IEEE Transactions on Education*, *68*(1), 103–116. https://doi.org/10.1109/TE.2024.3467912

Qiu, R., Sun, Z., & Yang, Y. (2022). DIMES: A Differentiable Meta Solver for Combinatorial Optimization Problems. In S. Koyejo, S. Mohamed, A. Agarwal, D. Belgrave, K. Cho, & A. Oh (Eds.), *Advances in Neural Information Processing Systems* (Vol.

35, pp. 25531–25546). Curran Associates, Inc. https://proceedings.neurips.cc/paper_files/paper/2022/file/a3a7387e49f4de290c23beea2dfcdc75-Paper-Conference.pdf

Radford, A., Narasimhan, K., Salimans, T., & Sutskever, I. (2018). *Improving Language Understanding by Generative Pre-Training*. https://cdn.openai.com/research-covers/language-unsupervised/language_understanding_paper.pdf

Rego, C., Gamboa, D., Glover, F., & Osterman, C. (2011). Traveling Salesman Problem Heuristics: Leading Methods, Implementations And Latest Advances. *European Journal of Operational Research*, *211*(3), 427–441. https://doi.org/10.1016/j.ejor.2010.09.010

Ruiz-Villafranca, S., Roldán-Gómez, J., Gómez, J. M. C., Carrillo-Mondéjar, J., & Martinez, J. L. (2024). A TabPFN-based Intrusion Detection System For The Industrial Internet of Things. *Journal of Supercomputing*, *80*(14), 20080–20117. https://doi.org/10.1007/s11227-024-06166-x

Saito, T., Otake, Y., & Wu, S. (2025). *Tabular foundation model for GEOAI benchmark problems BM/AirportSoilProperties/2/2025*. http://arxiv.org/abs/2509.03191

Saller, S., Koehler, J., & Karrenbauer, A. (2025). A survey on approximability of traveling salesman problems using the TSP-T3CO definition scheme. *Annals of Operations Research*, *351*(3), 2129–2190. https://doi.org/10.1007/s10479-025-06641-5

Schneider, J., Meske, C., & Kuss, P. (2024). Foundation Models. *Business and Information Systems Engineering*, *66*(2), 221–231. https://doi.org/10.1007/s12599-024-00851-0

Singh, S., Singh, A., Kapil, S., & Das, M. (2022). Utilization Of A TSP Solver For Generating Non-retractable, Direction Favouring Toolpath For Additive Manufacturing. *Additive Manufacturing*, *59*. https://doi.org/10.1016/j.addma.2022.103126

Skinderowicz, R. (2022). Improving Ant Colony Optimization Efficiency For Solving Large TSP Instances. *Applied Soft Computing*, *120*. https://doi.org/10.1016/j.asoc.2022.108653

Sun, Z., & Yang, Y. (2023). DIFUSCO: Graph-based Diffusion Solvers for Combinatorial Optimization. In A. Oh, T. Naumann, A. Globerson, K. Saenko, M. Hardt, & S. Levine (Eds.), *Advances in Neural Information Processing Systems* (Vol. 36, pp. 3706–3731). Curran Associates, Inc. https://proceedings.neurips.cc/paper_files/paper/2023/file/0ba520d93c3df592c83a611961314c98-Paper-Conference.pdf

Touvron, H., Lavril, T., Izacard, G., Martinet, X., Lachaux, M.-A., Lacroix, T., Rozière, B., Goyal, N., Hambro, E., Azhar, F., Rodriguez, A., Joulin, A., Grave, E., & Lample, G. (2023). *LLaMA: Open and Efficient Foundation Language Models*. http://arxiv.org/abs/2302.13971

Tu, X., He, Z., Huang, Y., Zhang, Z. H., Yang, M., & Zhao, J. (2024). An Overview Of Large AI Models And Their Applications. In *Visual Intelligence, 2(1)*. Springer. https://doi.org/10.1007/s44267-024-00065-8

Vesselinova, N., Steinert, R., Perez-Ramirez, D. F., & Boman, M. (2020). Learning Combinatorial Optimization on Graphs: A Survey with Applications to Networking. In *IEEE Access* (Vol. 8, pp. 120388–120416). Institute of Electrical and Electronics Engineers Inc. https://doi.org/10.1109/ACCESS.2020.3004964

Vu, N. G. H., Wang, K., & Wang, G. G. (2025). Effective Prompting With ChatGPT For Problem Formulation In Engineering Optimization. *Engineering Optimization*. https://doi.org/10.1080/0305215X.2025.2450686

Yang, H., Zhao, M., Yuan, L., Yu, Y., Li, Z., & Gu, M. (2023). Memory-efficient Transformer-based Network Model for Traveling Salesman Problem. *Neural Networks*, *161*, 589–597. https://doi.org/10.1016/j.neunet.2023.02.014

Zhao, C. S., & Wong, L. P. (2025). A Transformer-based Structure-aware Model For Tackling The Traveling Salesman Problem. *PLoS ONE*, *20*(4). https://doi.org/10.1371/journal.pone.0319711
22

Zheng, J., He, K., Zhou, J., Jin, Y., & Li, C. M. (2023). Reinforced Lin–Kernighan–Helsgaun Algorithms For The Traveling Salesman Problems. *Knowledge-Based Systems*, *260*. https://doi.org/10.1016/j.knosys.2022.110144

Zhou, C., Li, Q., Li, C., Yu, J., Liu, Y., Wang, G., Zhang, K., Ji, C., Yan, Q., He, L., Peng, H., Li, J., Wu, J., Liu, Z., Xie, P., Xiong, C., Pei, J., Yu, P. S., & Sun, L. (2024). A Comprehensive Survey On Pretrained Foundation Models: A History From BERT To ChatGPT. *International Journal of Machine Learning and Cybernetics*. https://doi.org/10.1007/s13042-024-02443-6

Zhou, C., Lin, X., Wang, Z., & Zhang, Q. (2025). *Learning to reduce search space for generalizable neural routing solver* arXiv. https://arxiv.org/abs/2503.03137